\documentclass[11pt,a4paper]{article}
\usepackage[hyperref]{acl2021}
\usepackage{times}
\usepackage{latexsym}

\usepackage{xcolor}
\usepackage[utf8]{inputenc}
\usepackage[english]{babel}

\usepackage{array, makecell}
\usepackage{amsmath,amsthm}
\usepackage{multirow}
\usepackage{microtype}
\usepackage{booktabs}
\usepackage{lipsum}  
\usepackage{subcaption}
\usepackage{adjustbox}
\usepackage{tikz}

\usepackage[noend]{algorithmic}
\usepackage{algorithm}

\usepackage{listings}

\colorlet{punct}{red!60!black}
\definecolor{background}{HTML}{EEEEEE}
\definecolor{delim}{RGB}{20,105,176}
\colorlet{numb}{magenta!60!black}

\lstdefinelanguage{json}{
    basicstyle=\scriptsize\ttfamily,
    numbers=left,
    numberstyle=\scriptsize,
    stepnumber=1,
    numbersep=8pt,
    showstringspaces=false,
    breaklines=true,
    frame=lines,
    backgroundcolor=\color{background},
    literate=
     *{0}{{{\color{numb}0}}}{1}
      {1}{{{\color{numb}1}}}{1}
      {2}{{{\color{numb}2}}}{1}
      {3}{{{\color{numb}3}}}{1}
      {4}{{{\color{numb}4}}}{1}
      {5}{{{\color{numb}5}}}{1}
      {6}{{{\color{numb}6}}}{1}
      {7}{{{\color{numb}7}}}{1}
      {8}{{{\color{numb}8}}}{1}
      {9}{{{\color{numb}9}}}{1}
      {:}{{{\color{punct}{:}}}}{1}
      {,}{{{\color{punct}{,}}}}{1}
      {\{}{{{\color{delim}{\{}}}}{1}
      {\}}{{{\color{delim}{\}}}}}{1}
      {[}{{{\color{delim}{[}}}}{1}
      {]}{{{\color{delim}{]}}}}{1},
}

\usepackage{microtype}

\aclfinalcopy % Uncomment this line for the final submission
\def\aclpaperid{2114} %  Enter the acl Paper ID here

\title{Learning to Generate Task-Specific Adapters from Task Description}

\author{Qinyuan Ye \quad Xiang Ren \\
  University of Southern California \\
  \texttt{\{qinyuany,xiangren\}@usc.edu} 
}

\begin{document}
\maketitle
\begin{abstract}
Pre-trained text-to-text transformers such as BART have achieved impressive performance across a range of NLP tasks. Recent study further shows that they can learn to generalize to novel tasks, by including task descriptions as part of the source sequence and training the model with (source, target) examples. At test time, these fine-tuned models can make inferences on new tasks using the new task descriptions as part of the input. 
% new task descriptions are included in the input and the model directly makes inference.
However, this approach has potential limitations, as the model learns to solve individual (source, target) examples (\textit{i.e.}, at the \textit{instance} level), instead of learning to solve tasks by taking all examples within a task as a whole (\textit{i.e.}, at the \textit{task} level). 
To this end, we introduce \textsc{Hypter}, a framework that improves text-to-text transformer's generalization ability to unseen tasks by training a \underline{hyp}ernetwork to generate task-specific, light-weight adap\underline{ter}s from task descriptions. 
Experiments on ZEST dataset and a synthetic SQuAD dataset demonstrate that \textsc{Hypter} improves upon fine-tuning baselines. Notably, when using BART-Large as the main network, \textsc{Hypter} brings 11.3\% comparative improvement on ZEST dataset.\footnote{Code and data can be found at \url{https://github.com/INK-USC/hypter}.}

\end{abstract}

\section{Introduction}

Pre-trained text-to-text models \cite{raffel2019exploring, lewis-etal-2020-bart} provide a unified formulation and off-the-shelf weights for a variety of NLP tasks, such as question answering \cite{khashabi-etal-2020-unifiedqa} and commonsense reasoning \cite{bosselut-etal-2019-comet}. In addition to their strong performance, text-to-text models naturally support generalizing to novel tasks, by incorporating task description as part of the source sequence and fine-tuning the model with (source, target) examples \cite{weller-etal-2020-learning}. At inference time, the model is required to perform unseen tasks with the source sequence containing new task descriptions.

\begin{figure}[t]
    \centering
\hspace*{-0.2cm}    
    \includegraphics[width=0.5\textwidth]{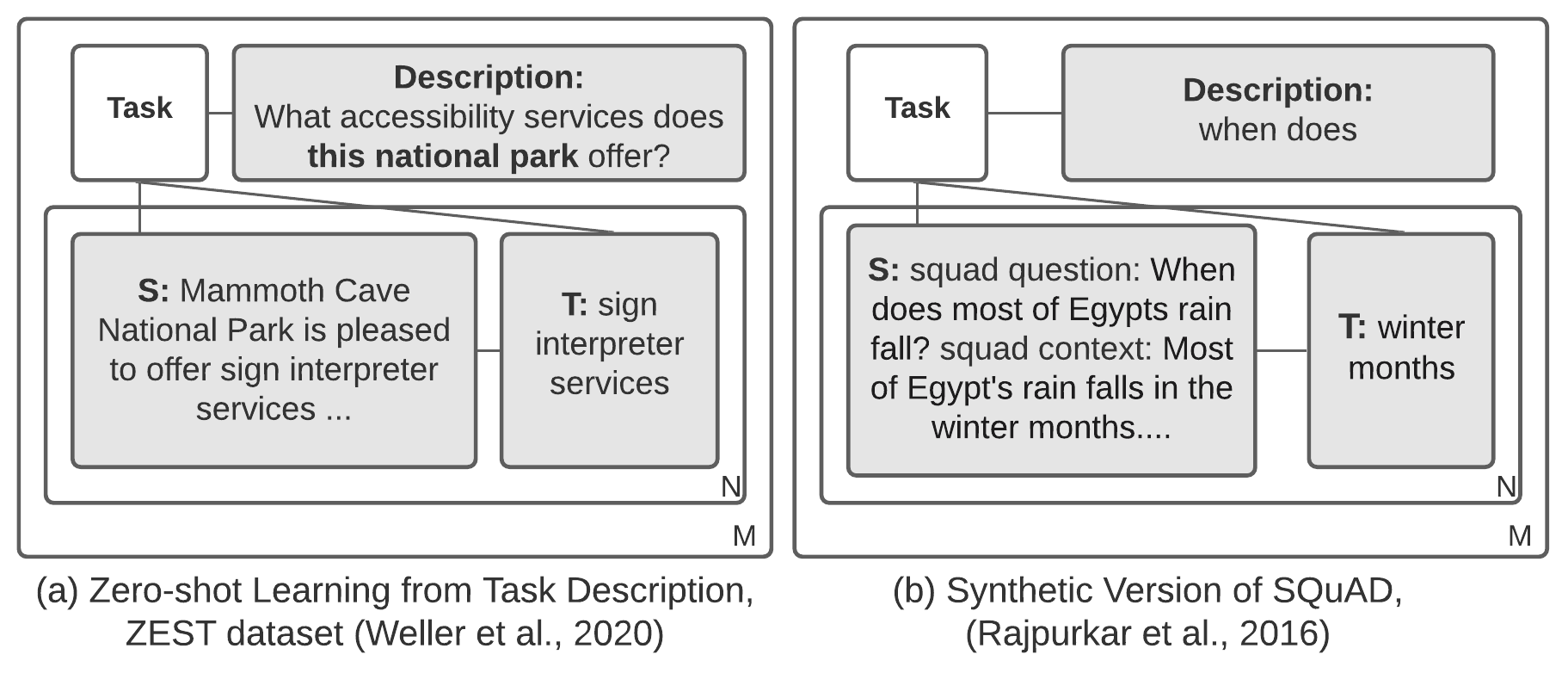}
    \vspace{-0.7cm}
    \caption{Instead of learning from (source, target) examples, in this paper we study the problem of \textit{learning from task descriptions} \cite{weller-etal-2020-learning}. The train set contains $M$ tasks, and the $i$-th task contains $N_i$ examples of $(s,t)$ pairs in text format. During test time, the learned model is required to directly make inferences on a new task given a task description.}
    \label{fig:intro}
    \vspace{-0.5cm}
\end{figure}
\begin{figure*}[t]
    \definecolor{mypink}{RGB}{249, 210, 222}
    \definecolor{mypurple}{RGB}{224, 211, 225}
    \definecolor{myblue}{RGB}{214, 235, 247}
    \centering
    \includegraphics[width=0.98\textwidth]{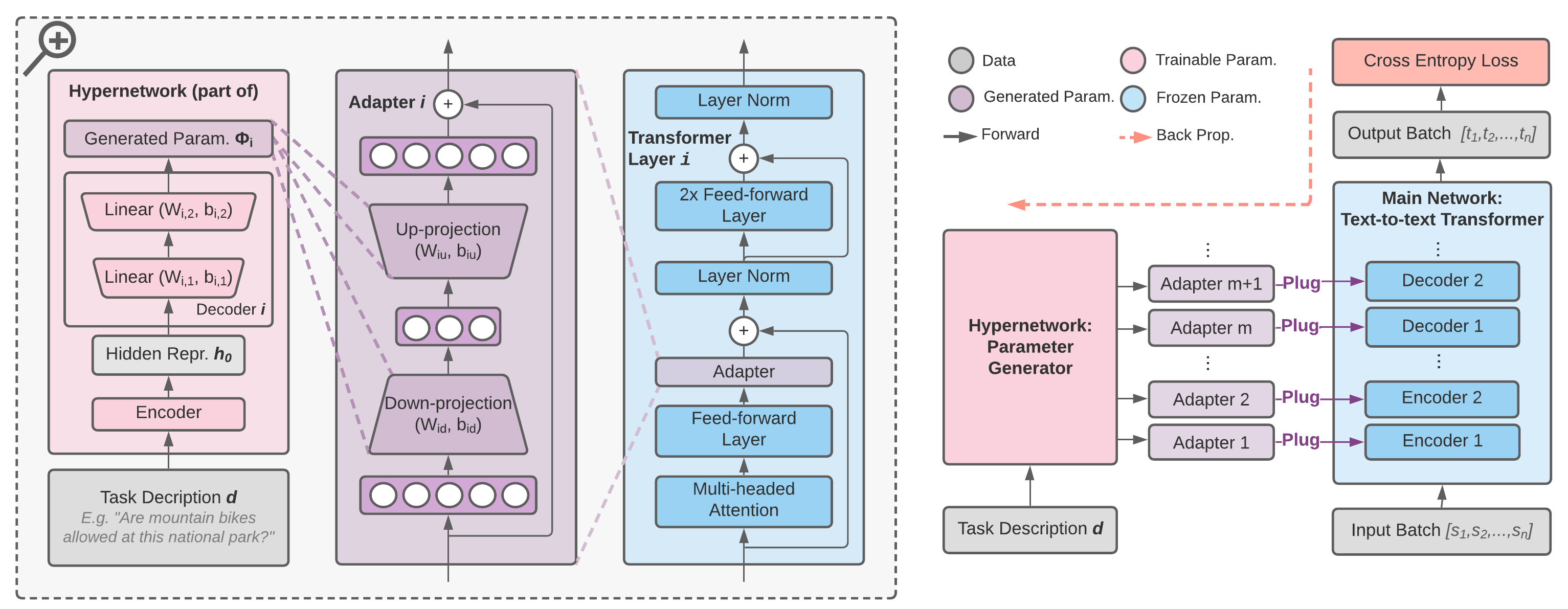}
    % \vspace{-0.3cm}
    \caption{\textbf{Illustration of \textsc{Hypter} Framework. Left:} {\setlength{\fboxsep}{0pt}\colorbox{mypink}{A hypernetwork}} generates parameter $\phi_i$ for {\setlength{\fboxsep}{0pt}\colorbox{mypurple}{task-specific adapter $i$}} that is plugged to {\setlength{\fboxsep}{0pt}\colorbox{myblue}{transformer layer $i$}} in the text-to-text model. \textbf{Right:} The adapted main network is evaluated on a task $(d, \mathcal{D})$. The final cross entropy loss is back-propagated to update the hypernetwork.}
    \label{fig:big}
    % \vspace{-0.3cm}
\end{figure*}
While this initial attempt shows positive results, there are two potential limitations for the direct fine-tuning approach. (1) Predictions can be sensitive to the task descriptions (or ``prompts'') that are heuristically designed \cite{jiang-etal-2020-know}. Paraphrasing the task description may lead to performance downgrade. (2) The model still learns from individual (source, target) examples, instead of learning to solve tasks at a higher level, by explicitly taking multiple examples within a task as a whole (see Fig. \ref{fig:intro}).
Meanwhile, applying existing zero-shot learning methods that supports task-level learning to text-to-text transformers is non-trivial. Methods designed specifically for classification problems, such as prototypical networks \cite{prototypical17snell}, cannot be directly applied to text-to-text models. Moreover, given the large size of text-to-text models, generating parameters for a whole model from the task description \cite{jin-etal-2020-language} is infeasible.

% In this paper, we aim to improve zero-shot learning ability of text-to-text transformer models by better incorporating task descriptions. Our study focuses on the problem of \textit{in-task} zero-shot learning, where \textit{e.g.}, inference on an unseen relation for slot filling; see Fig. \ref{fig:intro}), as opposed to \textit{cross-task} zero-shot learning (\textit{e.g.}, train a textual entailment model and test on question answering, \citealt{yin-etal-2020-universal}). 

% We propose a parameter generation module, that produces task-specific layer from the task description to adapt the model to a new task (i.e., \textit{adapter}).

In this work, we follow the settings in \cite{weller-etal-2020-learning} and aim to improve a model's generalization ability to unseen tasks by better incorporating task descriptions and using a task-level training procedure.
We introduce \textsc{Hypter}, a framework that employs a hypernetwork \cite{Ha2017HyperNetworks} to dynamically generate task-specific parameters (\textit{i.e.}, adapters) from task descriptions. Adapters \cite{pmlr-v97-houlsby19a} are light-weight modules that can be inserted into transformer layers for \textit{parameter-efficient} adaptation. 
Such formulation also effectively enables learning at the \textit{task} level, by learning to generate appropriate parameters for a task, and examine the model's competence on each task using multiple examples within that task.
This is in contrast to learning at the \textit{instance} level, by learning to generate the correct output for one specific input sequence. 
% Consequently our proposed training procedure will enable the learned hypernetwork to generalize at task level, supporting better generalization to unseen tasks.
% \xiang{it seems to me your approach has an explicit task representation (task-specific adapter) and prior work doesn’t --- so your learning of task is explicitly vs. prior work is implicit. Maybe just try to make this point clear?}

We apply \textsc{Hypter} to two datasets: ZEST \cite{weller-etal-2020-learning} and a synthetic version of SQuAD \cite{rajpurkar-etal-2016-squad}. We demonstrate that \textsc{Hypter} improves upon direct fine-tuning baselines. Notably, training with \textsc{Hypter} achieves 0.45\% absolute improvement (11.3\% comparative improvement) in Competence@90 metric on ZEST, when BART-Large is used as the main network.

\section{Problem Definition}
\vspace{-0.1cm}

We study the problem of \textit{learning from task description} \cite{weller-etal-2020-learning}, and aim to improve models' competence on unseen tasks at the inference time.
Formally, a task is denoted as a tuple of $(d, \mathcal{D})$, where $d$ is the natural language description of the task, and $\mathcal{D}=\{(s_1, t_1), ..., (s_n, t_n)\}$ contains (source, target) examples of this task (See Fig. \ref{fig:intro}). In our text-to-text formulation, both $s_i$ and $t_i$ are text sequences. At train time, both $d$ and $\mathcal{D}$ are available, while at test time, an unseen description $d$ is given, and the model is expected to predict the correct $t$ given input $s$ without further training. 

For instance, in the ZEST dataset \cite{weller-etal-2020-learning}, a train task description can be ``\textit{Are mountain bikes allowed at this national park?}'', while $\mathcal{D}$ contains twenty paragraphs for different national parks and twenty corresponding answers. During test time, a novel task may be ``\textit{Are there fish in this national park that live in caves?}'', and the model is asked to directly make inferences. 
% ZEST evaluates a model with four types of generalization: paraphrase, semantic flips, composition, and output structure.

% \input{figures_input/arch}

% Several existing datasets fall into the in-task zero-shot learning setting. Zero-shot text classification \cite{yin-etal-2019-benchmarking} introduces unseen classes at test time, while zero-shot slot filling \cite{levy-etal-2017-zero} presents unseen relations at test time. ZEST dataset \cite{weller-etal-2020-learning} contains new sub-tasks in testing portion, where the sub-task description is formatted as a generalized question.
% \footnote{Zest dataset \cite{weller-etal-2020-learning} formulate a ``task'' as a question. To avoid confusion, we refer to this as zero-shot QA.}. These benchmarks all fall into the \textit{in-task} category.

% \paragraph{Additional Notations for Adapters.} Our work is based on adapters \cite{pmlr-v97-houlsby19a}, light-weight modules that can be placed into transformer layers for parameter-efficient transfer learning. During learning, the main model is frozen, while only layer norm and adapter parameters are learnable. In this paper, we adopt a simplified design compared to the original paper (see Fig. \ref{fig:adapter})~---~In each transformer, exactly one adapter module will be added after the multi-headed attention. One adapter module contains two linear layers separated by an non-linearity activation layer. We denote $W_{id}, b_{id}$ as the parameter for down-projection for the adapter in layer $i$, and $W_{iu}, b_{iu}$ for the up-projection.
\section{Background: Adapters}
Our work is built on adapters \cite{pmlr-v97-houlsby19a}, light-weight modules that can be placed into transformer layers for parameter-efficient transfer learning. In the original paper, the main model is frozen during training, while only layer norm and adapter parameters are learnable. In this paper, we adopt a simplified design compared to the original paper (see Fig. \ref{fig:big} (Left))~--~In each transformer layer, exactly one adapter module will be added after the multi-headed attention. One adapter module contains two linear layers separated by an non-linearity activation layer. We use $(\mathbf{W}_{id}, \mathbf{b}_{id})$ to denote the down-projection parameters for the adapter in transformer layer $i$, and $(\mathbf{W}_{iu}, \mathbf{b}_{iu})$ for the up-projection parameters.

\section{Method}

\paragraph{Overview.} Fig. \ref{fig:big} provides an illustration of our \textsc{Hypter} framework. \textsc{Hypter} has two major parts: (1) A main network, which is a pre-trained text-to-text model. We instantiate the main network with BART-Base/Large \cite{lewis-etal-2020-bart}. (2) A hypernetwork, which generates adapters to be plugged into the main network. Fig. \ref{fig:big} (Left) contains a detailed illustration of how adapter parameters are generated and how adapter layers are incorporated into one transformer layer. 
% For additional information on adapters, see Appendix \ref{app:adapter}.

\paragraph{Hypernetwork.} 
The hypernetwork consists of an encoder and multiple decoders. 
The encoder maps the task description $d$ to a latent representation $\mathbf{h}_0$, while the decoders use $\mathbf{h}_0$ to generate adapter parameters $\phi$. In our work we instantiated the encoder with a RoBERTa-Base model \cite{liu2019roberta}, \textit{i.e.}, $\mathbf{h}_0 = \text{RoBERTa}(d)$.
For a text-to-text model with $n$ transformer layers, the hypernetwork contains $n$ decoders. Decoder $i$ uses $\mathbf{h}_0$ as input, and outputs adapter parameters $\phi_i$ for transformer layer $i$, \textit{i.e.}, 
$\mathbf{h}_{i,1} = \text{ReLU}(\mathbf{W}_{i,1}\mathbf{h_0} + \mathbf{b}_{i,1})$, $\mathbf{\phi}_i = \mathbf{W}_{i,2}\mathbf{h}_{i,1} + \mathbf{b}_{i,2}$.
% \begin{equation}\label{eq:phi-i-def}
% \begin{aligned}
%     \mathbf{h}_{i,1} &= \text{ReLU}(\mathbf{W}_{i,1}\mathbf{h_0} + \mathbf{b}_{i,1}),\\
%     \mathbf{\phi}_i &= \mathbf{W}_{i,2}\mathbf{h}_{i,1} + \mathbf{b}_{i,2}.
% \end{aligned}
% \end{equation}
Here $\mathbf{W}_{i,1}, \mathbf{b}_{i,1}, \mathbf{W}_{i,2}, \mathbf{b}_{i,2}$ are trainable parameters. The generated parameters $\mathbf{\phi}_i$ are sliced and reshaped to become parameters $[\mathbf{W}_{id}, \mathbf{b}_{id}, \mathbf{W}_{iu}, \mathbf{b}_{iu}]$ used in the adapter $i$.

\paragraph{Model Training.} We adopt a training schedule where we first train the main network, then train the hypernetwork while the main network is frozen. Conceptually, the first stage ensures that the main network captures the general ability across different tasks; the second stage allows the hypernetwork to learn to adapt the main network to a specific task. During the first stage the text-to-text model is fine-tuned with all $(\text{Concat}(d,s), t)$ examples in the training set. Here $\text{Concat}(d,s)$ means the concatenation of task description $d$ and input $s$. The learned main network from this stage also serves as the baseline method. 

During the second stage, we sample a task $(d, \mathcal{D})$ from the training set and sample a mini-batch of $(s,t)$ examples from $\mathcal{D}$. Given a description $d$, the hypernetwork generates adapter parameters $\phi_i$. We insert the resulting adapter layers into the main network, and compute the cross entropy loss $L$ of generating $t$ given input $\text{Concat}(d,s)$. The loss is end-to-end differentiable and is back-propagated to update the hypernetwork, while the main network is frozen. See Fig. \ref{fig:big} (Right) for illustration. This second stage of training effectively enables learning at the \textit{task} level. The loss $L$ characterizes the model's competence in the task $(d, \mathcal{D})$. Therefore, by optimizing $L$, the model is trained to \textit{solve tasks}.

% The overall training includes two stages. We first train a baseline model by directly fine-tuning the main model (see Sec. \ref{sec:baseline}). This stage ensures that the main model captures the general ability across different sub-tasks. We then freeze the main network and start training the hypernetwork, by sampling a task $(d, \mathcal{D})$ and sampling a batch of $(x,y)$ examples from $\mathcal{D}$. Given a description $d$, we first generate adapter parameters and apply them to the main network, then compute the cross entropy loss of generating $Y$ given input $\text{Concat}(X,d)$. The loss is end-to-end differentiable and is back-propagated to update the hypernetwork, meanwhile the main network is frozen.

\paragraph{Model Inference.} At test time the model is given an unseen task description $d$. The hypernetwork generates description-dependent adapter parameters, similar to the procedure during training. In this way, we obtain a model that is capable of making inferences for this new task.

\section{Experiments}

\subsection{Experiment Setup}
\paragraph{Datasets.} 
We use two datasets that fit our setup. The first one is Zero-shot Learning from Task Descriptions dataset (ZEST, \citealt{weller-etal-2020-learning}), which formulates task descriptions as generalized questions, and provides multiple source-target examples for each question. The performance is evaluated with a novel metric: ``Competence@K'', along with mean F1 score. 
Competence@K is the percentage of all tasks for which the model achieves mean F1 score higher than K.
For example, Competence@90=5 suggests that 5\% of all tasks can be solved with mean F1 better than 90\%. 
We report dev set performance, and hidden test set performance obtained from ZEST's official leaderboard.
% This dataset contains 538/114/599 sub-tasks in train/dev/test sets correspondingly.

% \qinyuan{Add more description about this (complained by Reviewer1).} # DONE
We construct the second dataset from SQuAD v1 \cite{rajpurkar-etal-2016-squad} to simulate the problem setting in this paper. We refer to this dataset as Synthetic SQuAD.
Specifically, we construct tasks from the original SQuAD train set according to ``question type'', the bi-gram containing the central question word (\textit{e.g.}, what, when). For example, ``when does'' questions are considered as a task, and ``what country'' questions are considered as another task. These bi-grams are used as ``task descriptions''. We select the 100 most frequent question types in SQuAD train set, and randomly subsample 64 examples from each type to formulate our dataset. We then randomly split the 100 types into 80/10/10 for train/dev/test. In addition, we select examples that fall into the 10 test question types from Natural Questions \cite{kwiatkowski-etal-2019-natural} and NewsQA \cite{trischler-etal-2017-newsqa}, and use these as out-of-domain test examples. Performance is evaluated with mean F1. We include the list of question types and more details about this dataset in Appendix~\ref{app:dataset}.

% We group the questions in SQuAD train set according to their ``question type'', the bi-gram containing the central question word. These bi-grams are also used as task descriptions. For example, popular question types in SQuAD include ``when does'', ``what country'', etc. We select the top 100 question types and subsample 64 examples from each type to formulate our synthetic dataset. We randomly split the 100 types into 80/10/10 for train/dev/test. We additionally select examples that fall into the 10 test question types from Natural Questions \cite{kwiatkowski-etal-2019-natural} and NewsQA \cite{trischler-etal-2017-newsqa}, and use these as out-of-domain test examples. Performance is evaluated with mean F1.

% \input{figures_input/adapter}
% \input{tables/input_output_format}

\paragraph{Baseline.}
To demonstrate the efficacy of the \textsc{Hypter} framework, we compare it to just its first half -- the main text-to-text transformer model that we obtain after the first stage of training. 
This is identical to the fine-tuning baseline method in \cite{weller-etal-2020-learning}, and there are no other applicable baselines to the best of our knowledge.

\paragraph{Training Details.} 
For each method, we train the model 7 times using different random seeds, and we report average and standard deviation.
We discuss other training details, including hyperparameters, in Appendix \ref{app:training}. Notably, we ensure all baseline models will not benefit from additional training, by tuning the number of epochs and using early stopping based on dev performance. This ensures the improvement brought by \textsc{Hypter} is not due to additional training.
% Notably, the two datasets are nascent and there are no other applicable baselines methods to the best of our knowledge. \label{sec:baseline}

% \paragraph{Training Details.} We use transformers \cite{wolf-etal-2020-transformers} for all our experiments. For hypernetwork training, we train up to 100 epochs (one epoch here refers an iteration over all sub-tasks). We update the hypernetwork every $b$ tasks, and we call $b$ as task batch size. When learning from one sub-task, we sample $b'$ examples within this task, and we call $b'$ as the example batch size. We select $b$ from \{4,8,16,32\}, $b'$ from \{4,8,16,32\}, adapter width $d$ from \{4,8,16,32\}, learning rate $\alpha$ from \{3e-6, 1e-5, 3e-5\}, based on dev set performance.

% \input{tables/zest}
\begin{table}[t]
\centering
\vspace{-0.0cm}
\scalebox{0.65}{
\begin{tabular}{llll}
\toprule
                    %   & \multicolumn{3}{c}{ZEST-Dev}  \\
Model                   & Mean-F1 & C@75   & C@90   \\
\midrule
Bart-Base & 28.44 ($\pm$1.58)  & 5.76 ($\pm$2.10)  & 0.74 ($\pm$0.00)   \\
\ \ + \textsc{Hypter}    & \textbf{28.96} ($\pm$1.15)  & \textbf{6.32} ($\pm$2.02){*}  & \textbf{1.08} ($\pm$0.62) \\
\midrule
Bart-Large (reported)  & 40     & 13     & 8     \\
Bart-Large & 41.17 ($\pm$1.16)  & 15.74 ($\pm$2.16) & 7.17 ($\pm$1.66)  \\
\ \ + \textsc{Hypter}    & \textbf{41.65} ($\pm$1.34) & \textbf{16.41} ($\pm$2.15){*} & \textbf{7.62} ($\pm$1.66){*}  \\
\bottomrule
\end{tabular}
}
\vspace{-0.1cm}
\caption{\textbf{Performance on ZEST Dev Set.} ``C@75/90'' refers to Competence@75/90 metric. We report mean and standard deviation over 7 runs. $*$ indicates statistical significance in a two-tailed paired t-test ($p<0.05$).}\label{tab:zest-dev}
\vspace{-0.1cm}
\end{table}
\begin{table}[t]
\centering
% \vspace{-0.2cm}
\scalebox{0.65}{
\begin{tabular}{llll}
\toprule
                    %   & \multicolumn{3}{c}{ZEST-Test}  \\
Model                   & Mean-F1 & C@75   & C@90   \\
\midrule
Bart-Base & 31.97 & \textbf{7.03}  & 2.23   \\
\ \ + \textsc{Hypter}    & \textbf{32.32} &  6.72 & \textbf{2.53}  \\
\midrule
Bart-Large (reported)  & 37.93  & 11.19  & 3.96     \\
Bart-Large & 40.13  & 10.91 & 3.98  \\
\ \ + \textsc{Hypter}    & \textbf{40.41}  & \textbf{11.35} & \textbf{4.43}  \\
\bottomrule
\end{tabular}
}
\vspace{-0.1cm}
\caption{\textbf{Performance on ZEST Test Set.} Performance obtained from ZEST official leaderboard\footnotemark[2].}\label{tab:zest-test}
\vspace{-0.5cm}
\end{table}

\subsection{Results}
\paragraph{Main Results.}
We present the results for ZEST in Table~\ref{tab:zest-dev}-\ref{tab:zest-test} and results for Synthetic SQuAD in Table \ref{tab:squad}. 
On ZEST test set, we observe that the Competence@90 metric is improved from 3.98 to 4.43 when using BART-Large, yielding an 11.3\% relative improvement. 
When BART-Base is used, C@90 is improved from 2.23 to 2.53. This demonstrates that by learning to solve tasks with \textsc{Hypter}, the model's generalization ability to unseen tasks is improved.
On Synthetic SQuAD dataset, we observe 0.74\% improvement with BART-Base and 0.41\% improvement with BART-Large. Additionally, models trained with \textsc{Hypter} achieves comparable or better performance on out-of-domain test sets, suggesting the learned task-solving ability is generalizable to new test distribution.\footnotetext[2]{{\url{https://leaderboard.allenai.org/zest/submissions/public}}}\setcounter{footnote}{2}\footnote{Unexpectedly, in Table \ref{tab:squad} we observe that performance of BART-Large on NewsQA is worse than that of BART-Base. We suspect that BART-Large may have overfit the SQuAD train set during the first stage of fine-tuning.}
It is a known issue that evaluating zero-shot performance can be tricky. We tried our best to reduce the randomness and instability by using different random seeds. In Table~\ref{tab:zest-dev} and Table~\ref{tab:squad}, we demonstrate that performance improvement is significant (p$<$0.05) in multiple settings, \textit{e.g.}, on ZEST dev set when C@75 metric is used.

\begin{table}[t]
\centering
\scalebox{0.65}{
\begin{tabular}{lllll}
\toprule
                    %   & \multicolumn{1}{c}{SQuAD-Syn-Dev} & \multicolumn{3}{c}{SQuAD-Syn-Test}  \\
Model                   & SQuAD   & NQ  & NewsQA  \\
\midrule
Bart-Base  & 74.79 ($\pm$0.91) & 49.78 ($\pm$0.95) & 56.37 ($\pm$0.90)  \\
\ \ + \textsc{Hypter}     & \textbf{75.53} ($\pm$0.68){*} & \textbf{50.39} ($\pm$1.01){*} & \textbf{56.41} ($\pm$0.85) \\
\midrule
Bart-Large & 79.32 ($\pm$0.34) & 59.21 ($\pm$0.89) & 55.41 ($\pm$0.54) \\
\ \ + \textsc{Hypter}  & \textbf{79.73} ($\pm$0.50) & \textbf{59.58} ($\pm$0.57) & \textbf{55.60} ($\pm$0.90) \\
\bottomrule
\end{tabular}
}
\caption{\textbf{Performance on Synthetic SQuAD dataset.} We report mean and standard deviation over 7 runs. NQ and NewsQA serve as out-of-domain test data.}\label{tab:squad}
% \vspace{-0.4cm}
\end{table}
\begin{figure}[t]
\centering
\begin{subfigure}{.23\textwidth}
  \centering
  \includegraphics[width=\textwidth]{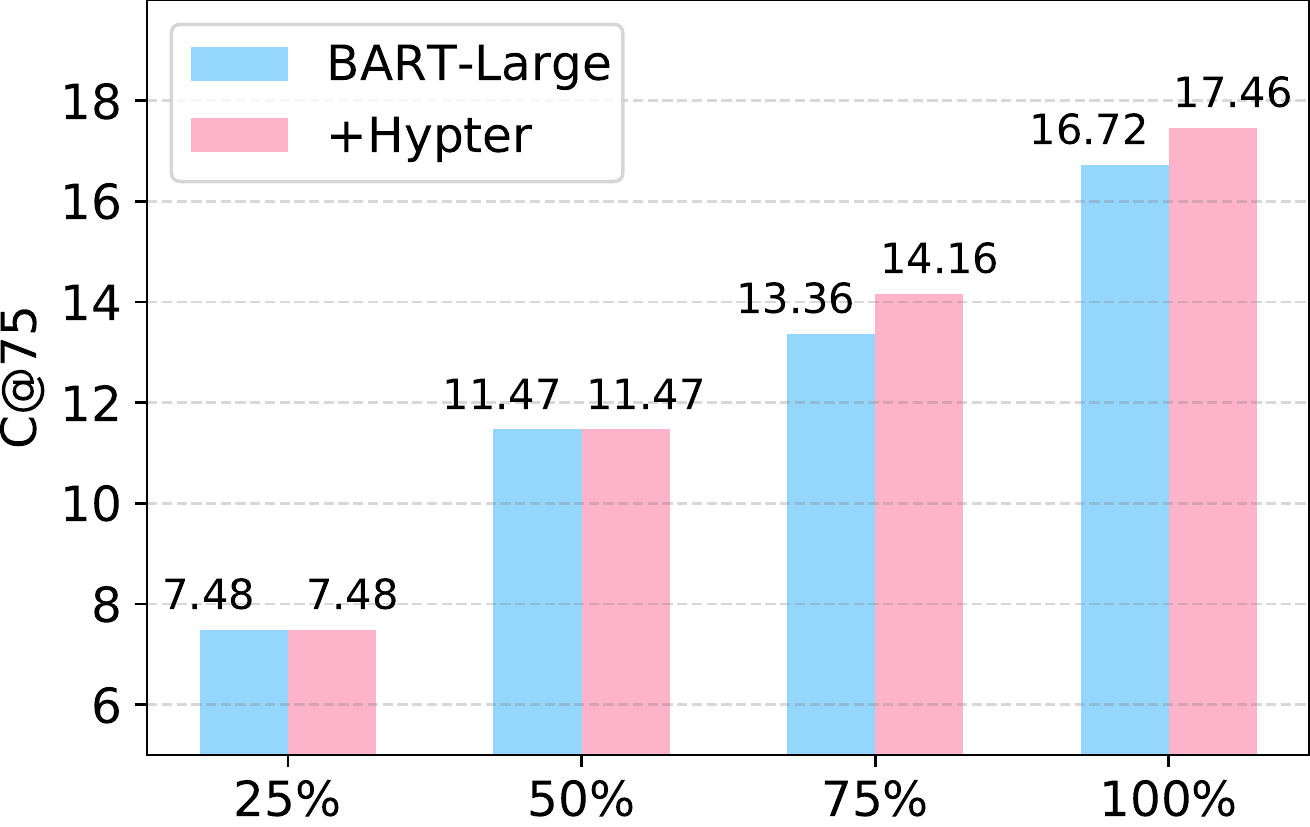}
  \caption{{\footnotesize\# Tasks}}
  \label{fig:sub1}
\end{subfigure}
\begin{subfigure}{.23\textwidth}
  \centering
  \includegraphics[width=\textwidth]{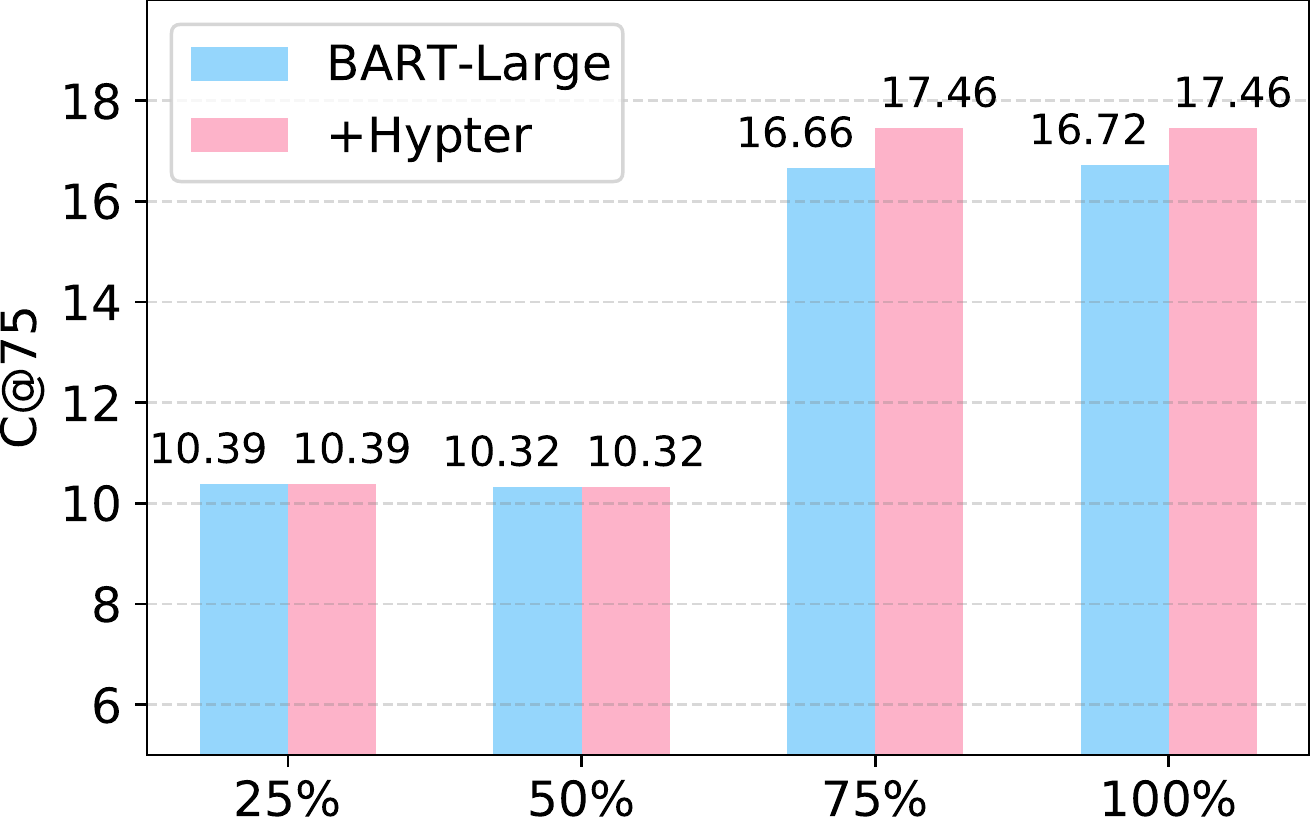}
  \caption{{\footnotesize\# Examples per Task}}
  \label{fig:sub2}
\end{subfigure}
\vspace{-0.3cm}
\caption{Competence@75 Performance on ZEST Dev when less training data is used.}
\vspace{-0.4cm}
\label{fig:ablation}
\end{figure}

% We suspect BART-Base's capacity is too small for the challenging task of ZEST and thus the improvement brought by \textsc{Hypter} is less significant.
% On ZSRE dataset, we also observe that \textsc{Hypter} brings improvement to the fine-tuning baseline. In addition, we observe that: (1) By comparing the two categories of with/without description, we conclude that including informative description of relations in the prompt improves performance. (2) The improvement brought by \textsc{Hypter} is less significant, or sometimes marginal, on ZSRE dataset. We suspect this is due to ZSRE dataset's small size (84 sub-tasks in train set compared to 553 in ZEST dataset); this creates additional challenges for training the hypernetwork.

\paragraph{Model Behavior Analysis on ZEST.} ZEST dataset provides a comprehensive analysis protocol by splitting tasks into different generalization types (base, paraphrase, composition, semantic flips, and output structure) and defining four error types (recall, precision, partial, and other). Compared to the BART-Large fine-tuning baseline, our model achieves better performance in ``base'' and ``paraphrase'' categories in the ZEST official test set. We also manually inspected dev set predictions produced by the baseline and our model. We found the predictions corrected by our method span across the four error types. In particular, the proposed method flipped two ``n/a'' predictions into the correct answers in the task ``Which royalty was this dog breed popular with?'' (``base'' category), reducing the recall errors and improving the competence metric. We do not observe more granular model behavioral patterns beyond this point.

\paragraph{Study of Data Efficiency.} We study whether \textsc{Hypter} is effective when trained with (1) fewer tasks, while the number of examples per task is unchanged; (2) fewer examples per task, while the number of total tasks is kept constant. We experiment with ZEST and BART-Large, and show the performance in Fig. \ref{fig:ablation}. We observe that \textsc{Hypter} is effective when trained with 75\%/100\% tasks, but does not improve performance with fewer tasks. This is reasonable since \textsc{Hypter} learns at the \textit{task} level (taking one task as an ``example''), and 50\% of the tasks may be insufficient. We also observe performance improvement with 75\%/100\% examples per task, but not with fewer examples. This suggests sufficient number of examples per task is necessary for \textsc{Hypter} to generate effective adapters.

% \paragraph{Significance Test.}
% One may achieve better generalization ability to unseen tasks with larger pre-trained models with billions of parameters. However this requires advanced computation resources such as TPUs. In this case, we consider \textsc{Hypter} as an alternative by augmenting a medium-sized pre-trained model with a hypernetwork, with all our experiments in this work being able to be fitted into one GPU with 48GB memory.
% So far we've demonstrated \textsc{Hypter}'s efficacy on two datasets. We 

% have several observations. Firstly, incorporating relation description as prompt improves performance by providing additional information. Secondly, additional training with \textsc{Hypter} brings 

% \subsection{Further Analysis}
% \paragraph{Ablation.} freeze/unfreeze roberta (encoder in the hypernetwork); use released bart/use fine-tuned bart.
% \paragraph{Other Attempts We've Made.} use 50\% to fine-tune, then use hypter.
% \input{sections/6_experiments_summarization}
\section{Related Work}

\paragraph{Zero-shot Learning with Transformers.} Zero-shot learning (ZSL) has been explored for various NLP tasks, including text classification \cite{yin-etal-2019-benchmarking}, entity linking \cite{logeswaran-etal-2019-zero} and entity typing \cite{obeidat-etal-2019-description}. Several works study cross-task transfer by unifying the input-output format, \textit{e.g.}, relation extraction as machine reading comprehension \cite{levy-etal-2017-zero}, named entity recognition as machine reading comprehension \cite{li-etal-2020-unified}. Such formulation allows generalization to unseen relation or named entity types at test time. 
Learning from task descriptions \cite{weller-etal-2020-learning} and instructions \cite{Mishra2021NaturalIB} can be considered as a sub-category in zero-shot learning, with the goal of generalizing to unseen tasks during inference.
% A separate line of work studies ZSL in cross-lingual settings, with unseen languages at test time \cite{nooralahzadeh-etal-2020-zero}.

\paragraph{Adapters for Transformers.} \citet{pmlr-v97-houlsby19a} proposed adapter layers for parameter-efficient transfer learning in NLP. Adapter layers, which adopt a bottleneck architecture with two linear layers, are added after each multi-headed attention layer and each feed-foward layer in a pre-trained transformer. 
Adapters have been recently applied to multi-lingual settings, with successes in NER, QA and commonsense reasoning \cite{pfeiffer-etal-2020-mad,philip-etal-2020-monolingual,artetxe-etal-2020-cross}.

\paragraph{Hypernetworks and Contextual Parameter Generators.} Hypernetwork \cite{Ha2017HyperNetworks} is a broad concept of ``using one network to generate the weights for another network''. This concept has been broadly applied to visual reasoning \cite{perez2018film}, zero-shot image classification \cite{jin-etal-2020-language}, etc. 
% \citet{platanios-etal-2018-contextual} introduced contextual parameter generator (CPG) which extends from hypernetworks by conditioning on ``well-defined context based on the input data''. 
Closely related to our work, UDapter \cite{ustun-etal-2020-udapter} studies multilingual dependency parsing by generating adapter parameters. Our work is more generalizable as we do not restrict task format (dependency parsing v.s. general text-to-text tasks) or relations between sub-tasks (cross-lingual tasks v.s. tasks with text-form descriptions).
\section{Conclusion}
In this paper, we introduced \textsc{Hypter}, a framework to improve text-to-text transformer's generalization ability to unseen tasks. \textsc{Hypter} enhances task-specific abilities by inserting adapters generated with a hypernetwork, meanwhile it maintains the model's general task-solving ability by freezing main model parameters. We demonstrated the effectiveness of \textsc{Hypter} on two datasets. Future work may explore teaching models with compositional instructions using \textsc{Hypter}, or propose robust fine-tuning methods that help the model generalize to unseen data. It is also necessary to construct a large dataset of diverse NLP tasks to facilitate future research in this direction.

% In the future, we will include more empirical analysis of the proposed approach, including joint training of the main network and the hypernetwork; ablation study of our design choices. We will also explore the more challenging problem of cross-task zero-shot learning.

\section*{Acknowledgments}
This research is supported in part by the Office of the Director of National Intelligence (ODNI), Intelligence Advanced Research Projects Activity (IARPA), via Contract No. 2019-19051600007, the DARPA MCS program under Contract No. N660011924033 with the United States Office Of Naval Research, the Defense Advanced Research Projects Agency with award W911NF-19-20271, and NSF SMA 18-29268. The views and conclusions contained herein are those of the authors and should not be interpreted as necessarily representing the official policies, either expressed or implied, of ODNI, IARPA, or the U.S. Government. 
We would like to thank anonymous reviewers and collaborators in USC INK research lab for their constructive feedback.

\bibliography{anthology,acl2021}
\bibliographystyle{acl_natbib}

\clearpage
\appendix
% \section{Adapters}
% \label{app:adapter}
% Our work is built on adapters \cite{pmlr-v97-houlsby19a}, light-weight modules that can be placed into transformer layers for parameter-efficient transfer learning. During learning, the main model is frozen, while only layer norm and adapter parameters are learnable. In this paper, we adopt a simplified design compared to the original paper (see Fig. \ref{fig:big} (Left))~--~In each transformer, exactly one adapter module will be added after the multi-headed attention. One adapter module contains two linear layers separated by an non-linearity activation layer. We denote $(\mathbf{W}_{id}, \mathbf{b}_{id})$ as the parameter for down-projection for the adapter in layer $i$, and $(\mathbf{W}_{iu}, \mathbf{b}_{iu})$ for the up-projection.

\section{Dataset Details}
\label{app:dataset}
\paragraph{ZEST.} ZEST dataset is released at \url{https://ai2-datasets.s3-us-west-2.amazonaws.com/zest/zest.zip}. ZEST leaderboard is hosted at \url{https://leaderboard.allenai.org/zest/submissions/public}.
\paragraph{Synthetic SQuAD.} We build our synthetic dataset from the processed version of SQuAD, Natural Questions and NewsQA in MRQA Shared Task 2019 \cite{fisch2019mrqa} (\url{https://mrqa.github.io/2019/}). We provide the script to reconstruct the data we use in our released code. We list the bi-grams we use to formulate synthetic tasks and their train/dev/test partition in Listing 1.

% \vspace{-0.1cm}
\begin{lstlisting}[language=json,firstnumber=1,caption={Train/Dev/Test Partition in Synthetic SQuAD dataset.}]
"train": ["why were", "what years", "who said", "what percent", "when did", "where do", "who is", "how are", "what decade", "how does", "how long", "where was", "what has", "which two", "who was", "who were", "where are", "where does", "what did", "how far", "what organization", "what does", "what group", "what would", "how did", "who has", "who created", "how many", "what name", "what types", "what two", "which city", "who are", "how is", "what event", "what are", "what century", "what area", "whom did", "why was", "who wrote", "why are", "where is", "how old", "when is", "what caused", "who did", "where did", "what happened", "what state", "what kind", "what time", "what famous", "what's the", "what day", "what is", "what company", "what were", "why do", "what new", "what date", "what do", "what color", "which group", "what country", "how can", "what have", "where can", "what period", "which year", "when was", "what other", "what happens", "was the", "what was", "which of", "when were", "what sort", "what city", "what year"],
"dev": ["what month", "why is", "what part", "what term", "how was", "how were", "how do", "who led", "which country", "when does"],
"test": ["where were", "what political", "what religion", "why did", "what type", "what language", "who had", "what percentage", "what can", "how much"]
\end{lstlisting}
\vspace{-0.2cm}

\section{Training Details} 
\label{app:training}
We use transformers \cite{wolf-etal-2020-transformers} for all our experiments. All experiments are done with one single GPU. We use NVIDIA Quadro RTX 8000, NVIDIA Quadro RTX 6000, or NVIDIA GeForce RTX 2080 Ti, depending on availability.

For text-to-text model fine-tuning, we select learning rate from \{1e-5, 3e-5, 5e-5\}, and select the total number of epochs from \{5, 10, 15, 20, 30\} for ZEST and \{10, 20, 30, 50, 100\} for synthetic SQuAD. We use a fixed batch size of 32.

For hypernetwork training, we train up to 100 epochs (one epoch here refers to an iteration over all tasks). We update the hypernetwork every $b$ tasks, and we call $b$ as task batch size. When learning from one task, we sample $b'$ examples within this task, and we call $b'$ as the example batch size. We greedily and sequentially select adapter width $d$ from \{4,8,16,32\}, learning rate $\alpha$ from \{3e-6, 1e-5, 3e-5, 1e-4\}, $b$ from \{4,8,16,32\}, $b'$ from \{4,8,16,32\}, based on dev set performance.

\section{Additional Baseline}
Another reasonable baseline is to fine-tune a text-to-text model together with randomly initialized adapters plugged in it. We experiment with this method using BART-Large and list the performance in Table~\ref{tab:zest-ftwadapter}. We do not observe significant differences between the two methods (p=0.8840 for C@75, p=0.8118 for C@90 in two-tailed paired t-test). 
% On a side note, we choose paired t-test because for each random seed, hypernetwork training (stage 2) is dependent on the fine-tuned model obtained in stage 1.
\begin{table}[h]
\centering
\scalebox{0.56}{
\begin{tabular}{llll}
\toprule
Model                   & Mean-F1 & C@75   & C@90   \\
\midrule
Bart-Large & 41.17 ($\pm$1.16)  & 15.74 ($\pm$2.16) & 7.17 ($\pm$1.66)  \\
Bart-Large with Adapters &  39.76 ($\pm$1.26)  & 15.61 ($\pm$1.14) & 6.96 ($\pm$1.15)\\
\bottomrule
\end{tabular}
}
\caption{Performance comparison when adapters are plugged / not plugged during fine-tuning.}\label{tab:zest-ftwadapter}
\vspace{-0.5cm}
\end{table}

\section{Dev Set Performance of Models Submitted to ZEST Leaderboard}
In Table~\ref{tab:zest-submitted} we present the dev performance of models submitted to the leaderboard. The submitted models are the ``first-runs'' in the 7-run series, as we add the 7-run experiments and significance test later on, following a reviewer's suggestion.
\begin{table}[h]
\centering
% \vspace{-0.2cm}
\scalebox{0.59}{
\begin{tabular}{lcccccccccccc}
\toprule
Model                   & Mean-F1 & C@75   & C@90   \\
\midrule
Bart-Base & 29.72  & 7.87  & \textbf{4.05}   \\
\ \ + \textsc{Hypter}    & \textbf{29.81}  &  \textbf{8.67} & \textbf{4.05} \\
\midrule
Bart-Large (reported)  & 40     & 13     & 8    \\
Bart-Large & 42.10  & 16.72  & 8.85  \\
\ \ + \textsc{Hypter}    & \textbf{43.50}  & \textbf{17.46} & \textbf{9.64} \\
\bottomrule
\end{tabular}
}
% \vspace{-0.2cm}
\caption{Dev set performance of models submitted to ZEST leaderboard.}\label{tab:zest-submitted}
\vspace{-0.5cm}
\end{table}

\section{Discussion}
It is worth noting that the efficacy of \textsc{Hypter} is at the cost of introducing new parameters in the hypernetwork. To generate adapter parameters, more parameters are introduced and trained in the hypernetwork. One may achieve better generalization ability to unseen tasks with larger pre-trained models with billions of parameters. In this case, we consider \textsc{Hypter} as an alternative by augmenting a medium-sized pre-trained model with a hypernetwork. Meanwhile, we highlight our contribution to be the concept of generating task-specific adapters from descriptions and \textsc{Hypter}'s task-level training procedure.

\end{document}